\documentclass[review]{elsarticle}

\usepackage{lineno,hyperref}
\usepackage{graphicx}
\usepackage{subfig}
\usepackage[linesnumbered,ruled,lined]{algorithm2e}
\usepackage{float}
\usepackage{tikz}
\usepackage{epstopdf}
\usepackage{amssymb}
\usepackage{amsmath}
\usepackage{booktabs}
\usepackage{array}
\usepackage{multirow}
\usetikzlibrary{arrows.meta}
\usepackage{natbib}
\usepackage{cleveref}

\journal{X. X. X}

\begin{document}

\begin{frontmatter}
		
		\title{Heat Conduction Plate Layout Optimization using Physics-driven Convolutional Neural Networks}
		
		\author[A]{Hao Ma}
		\ead{hao.ma@tum.de}
		
		\author[A,B]{Yang Sun \corref{mycorrespondingauthor}}
		\cortext[mycorrespondingauthor]{Corresponding author. Tel.: +49 89 289 16492.}
		\ead{sunyang-007@hotmail.com}
			
		\author[B]{Mario Chiarelli}
		\ead{mario.rosario.chiarelli@unipi.it}

		\address[A]{TUM School of Engineering and Design, Technical University of Munich, 85748 Garching, Germany}
		\address[B]{Department of Civil and Industrial Engineering, University of Pisa, Italy}
		
\begin{abstract}
The layout optimization of the heat conduction is essential during design in engineering, especially for thermal sensible products.
When the optimization algorithm iteratively evaluates different loading cases, the traditional numerical simulation methods used usually lead to a substantial computational cost. 
To effectively reduce the computational effort, data-driven approaches are used to train a surrogate model as a mapping between the prescribed external loads and various geometry.
However, the existing model are trained by data-driven methods which requires intensive training samples that from numerical simulations and not really effectively solve the problem.   
Choosing the steady heat conduction problems as examples, this paper proposes a \emph{Physics-driven Convolutional Neural Networks} (PD-CNN) method to infer the physical field solutions for random varied loading cases.
After that, the \emph{Particle Swarm Optimization} (PSO) algorithm is used to optimize the sizes and the positions of the hole masks in the prescribed design domain, and the average temperature value of the entire heat conduction field is minimized, and the goal of minimizing heat transfer is achieved.
Compared with the existing data-driven approaches, the proposed PD-CNN optimization framework not only predict field solutions that are highly consistent with conventional simulation results, but also generate the solution space with without any pre-obtained training data. 
\end{abstract}


\end{frontmatter}

\clearpage

\section{Introduction}\label{sec:Introduction}

Layout optimization is a cutting-edge research direction in the field of structural optimization and of great significance in the initial design stage of the product \cite{bendsoe2003topology}. 
As a conceptual design, it provides the optimal geometry of the structure and  usually carried out simultaneously with the shape and size optimization. 
For the products made by high thermal conductivity materials, the layout affects the heat conduction performance significantly and further the operation of components and systems. 
So, the objective of layout optimization is to establish an available geometry that provide a good heat distribution of the domain and thermal environment \cite{chen2016optimization}.
Although the layout optimization has a great potential to improve the geometry design for a precised heat loading, the process is time-consuming in some practical engineering problems.
The alternative layouts must be represented with acceptable resolutions. 
After that, numerical simulation approaches, such as \emph{Finite Element Method} (FEM) \cite{hughes2012finite} and \emph{Finite Volume Method} (FVM) \cite{gu2017application}, are used to evaluate the temperature response of different layout schemes. 
Although these numerical methods can provide correct evaluations for optimization work, they are usually time-consuming, especially for high-precision, very fine meshes.
Because of the intensive computational effort involved in every single layout case, the optimization loop is a challenge task when directly integrating the simulation results from the numerical analysis tool \cite{chen2020heat}.

From mathematical point of view, one possible approach to get rid of this severe computational burden is to utilize an explicit mapping between the parameters characterizing the prescribed structural geometry/layout and those responding temperature response \cite{lei2019machine}.
With abundant training methods and high-performance computing resources, \emph{Machine Learning} (ML) has been applied for many scientific research fields, such as fluid mechanics \cite{brunton2020machine,ma2020supervised,ma2021generative}, heat transfer \cite{farimani2017deep}, and other key areas \cite{tanaka2021deep,bourilkov2019machine}.
Among all of these, ML, especially deep learning, has been used to train a surrogate model to establish the mapping.
After the training process, the surrogate model extracts the complicated relationship between input parameters and output solutions, and is able to make the accurate predictions instantly according to the precised conditions  \cite{anzai2012pattern,sosnovik2019neural,muller2020surrogate}.
For example, Thuerey et al. proposed a deep learning framework based on the U-net architecture for inferring the \emph{Reynolds average Navier-Stokes} (RANS) solution and proved that it is much faster than the traditional \emph{Computational Fluid Mechinics} (CFD) solver \cite{thuerey2020deep}. 
Based on the this framework, Ma et al. used \emph{Convolutional Neural Networks} (CNN) to study the mixing characteristics of cooling film in a rocket combustion chamber \cite{ma2020supervised}, and later proposed a novel framework with physical evaluators, that is, to predict spray-based solutions by generating adversarial networks \cite{ma2021generative}. 
Bhatnagar et al. developed the CNN method for aerodynamic flow fields inference, while others studied the predictability of laminar flow  \cite{bhatnagar2019prediction}. 
Chen et al. used a graph neural network to predict transonic flow \cite{chen2019u}. 

Except the research using ML to train surrogate models for predicting accurate physical solutions, there are also a few works referring to optimization tasks \cite{chen2020heat,tao2019application,del2019deep}.
Eisman et al. used data-driven Bayesian methods to conduct optimization for generating object shapes with improved drag coefficients \cite{eismann2017shape}.
Li et al. \cite{li2020efficient} proposed a efficient framework based on \emph{Generative Adversarial Network} (GAN) for aerodynamic geometric shape  optimization.  
Chen et al. trained a neural networks model to infer the flow field and then used it as a surrogate model to perform airfoil shape optimization problems based on the gradient descent method.
It is proved that the optimization framework based on DNN can solve general aerodynamic design problems \cite{Chen_2021}.
Wang et al. developed a conditional generative adversarial neural network model to build the mapping between the input parameters and the surface temperature distribution. 
And then used the sparrow search algorithm to adjust and search the optimal parameters of components \cite{wang2022optimization}.

The work described above shows the ability of deep learning model to predict accurate physical solutions according to the precised conditions and presents the great potential in optimization tasks.  
However, these methods are based on data-driven and also require expensive training sample acquisition costs (numerical simulations or experimental results), which do not truly address the need to reduce computational burden \cite{ma2021physics}.
Actually, there is also a type of physics-driven method which require no or few training data. 
Mazier Rassi et al. introduced the laws of physics into machine learning algorithms and named it \emph{Physics Informed Neural Network} (PINN) \cite{raissi2017physics}. 
Introducing the \emph{Partial Differential Equations} (PDEs) into the loss function, the neural networks is trained to meet the corresponding PDEs and boundary conditions and eventually can predict a solution that satisfies physics. 
Compared with the data-driven methods, it eliminates the dependence of machine learning on training data and significantly reduces the cost of data set generation \cite{zhu2019physics,geneva2020modeling}. 
In addition, Ma et al. firstly proposed a combined method that both data and physics guide CNN to predict the solution of the temperature field \cite{ma2020combined}.
Different from the conventional data-driven methods of machine learning, a large amount of input data is not essential for physics-driven methods, which significantly improves the practical applicability and feasibility when the data are obtained  difficultly \cite{lu2019deepxde,sun2020surrogate}.  

In this research, based on the \emph{Physics-driven Convolutional Neural Networks} (PD-CNN), the prediction of the temperature field of the steady-state heat conduction plate is realized.
Also, an optimization framework based on \emph{Particle Swarm Optimization} (PSO) algorithm was established to search the optimal  size and position of the thermal insulation hole. 
As far as we know, this is the first time a physics-driven method is used to train a surrogate model for an optimization work.
The paper mainly consists of three parts, physics-driven training of the surrogate model, temperature field prediction, and the layout optimization using PSO algorithm. 
After the introduction, the preliminaries, including the U-net architecture and the physical loss function, are described in Section 2. 
Then the training process of the surrogate model and the prediction accuracy under the different heat loads and layouts are discussed in Section 3. 
The optimization results are presented and analyzed in Section 4. Finally, a conclusion and outlook are given in Section 5.

\section{Methodology}\label{sec:Methodology}


In this section, the U-net architecture of CNN used to predict the physical field are described. 
Then the physics-driven method using Laplace equation as the loss function is introduced. At last, the layout optimization using \emph{Particle Swarm Optimization} (PSO) algorithm is represented.

\subsection{U-net CNN for Temperature Prediction}
The U-net architecture is firstly proposed for computer vision for biomedical image segmentation and it is a convolutional encoder-decoder deep regression framework \cite{ronneberger2015u}. The encoder is used to capture context while the decoder is used to make high resolution pixel wise predictions, which are named as contracting path and expanding path, respectively. 
Particularly, a skip connection structure is also used in this architecture to combine the features from the contracting path with the upsampled feature from the corresponding expanding path. 

\begin{figure}[!h]
	\centering
	\includegraphics[width=1\textwidth]{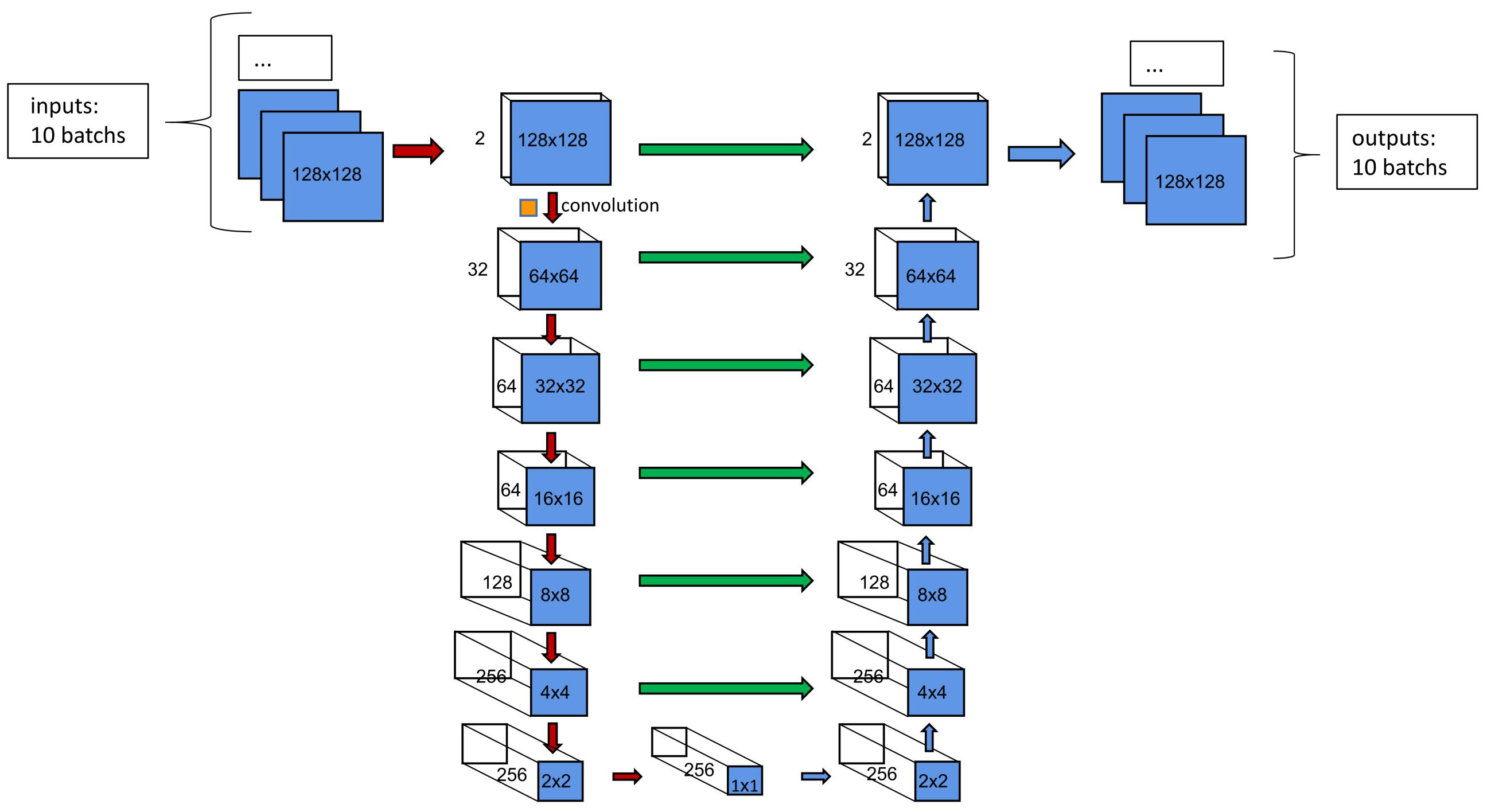}
	\caption{Schematic of the CNN architecture. 
		Each blue box represents a multi-channel feature map
		generated by the last convolutional layer. 
		The generating matrix size are denoted in the box.
		Green arrows denote ``skip-connections" via concatenation.}
	\label{figs:U-net}
\end{figure}

As shown in \Cref{figs:U-net}, the architecture in this paper has 17 layers corresponding to convolutional blocks. 
Each convolutional block is defined by batch normalization, active function, convolutional calculation, and dropout \cite{paszke2019pytorch,yamashita2018convolutional}. 
There are two inputs of the architecture on the left side, the first one is the initial temperature field with boundary conditions and the heat source is applied on the left side of the area.
The second one is the mask which defines the layout, where inside the rectangular holes the values of mask are zero and outside are one. 
So, the inputs are defined by a tensor with four dimensions: (10, 2, 128, 128). 
The first dimension is the number of the temperature field cases in one batch \cite{ruder2016overview}. 
The second one is the channels which represents the initial or the geometry masks. 
The last two dimensions are expressed as square matrices with dimensions $128 \times 128$. 
The input data is normalized to accelerate the training of deep networks.

The U-net architecture used in this paper is symmetrical, which means the encoding and decoding processes have the same depth, the corresponding blocks are identify. 
The left side of the architecture is the encoding process in which the values of the temperature field are progressively down-sampled using convolutional calculations. 
The encoding process is used for recognizing the geometry and the boundary conditions of the input physical field, to extract the necessary features using convolution operation layer by layer. 
The square filters are applied to conduct the convolutional calculation layer by layer until the matrices with only one data point are obtained (see the block at the vertex of the triangle structure). 
In this process, the large-scale information can be extracted as the feature channels, which increase through layers. 
For the decoding process, which corresponds to the right side of the structure, can be regarded as an inverse convolutional process and mirrors the operations of the encoding part. 
By decreasing the feature channel amounts and increasing spatial resolution, the solutions of the temperature field are reconstructed in the up-sampling layers. 
The horizontal arrows and vertical arrows, as shown in Figure \ref{figs:U-net}, concatenate the encoding blocks and the corresponding decoding blocks \cite{ronneberger2015u}. 
The motivation of these connections is doubling the feature channels of the decoding block, which are connected by green horizontal arrows, and making the neural networks consider the information from the encoding layers that are connected by red and blue vertical arrows.

To train the CNNs, stochastic gradient descent optimization is used, which needs a loss function to calculate the difference between the results and the ground truth. Furthermore, in each epoch, the weights of every convolutional block are updated with Adam optimizer during the back propagation process \cite{kingma2014adam}.

\subsection{Physics-driven Training Approach} 
Recently, the physics-driven training methods for deep learning have shown particular promise for the physical fields prediction work for their advantages that reduce the heavy requirement of the large amount of training data. 
In the typical data-driven method, the mathematical formulation of the loss function compares the difference between prediction result and the ground truth as: 
\begin{equation}\label{con:Loss}
L_{data}=|T_{out}-T_{truth}|
\end{equation}

As shown in \Cref{con:Loss}, the subscript "data" denotes data-driven, the $T_{out}$, $T_{truth}$ are output and true temperature fields, respectively. 
While in this paper, the loss function is built based on the Fourier's law.
The physics law of the heat conduction can be expressed with a second-order PDE, the Laplace equation. 
The system assumes to be no inner heat source contribution, and the thermal conductivity is considered constant. 
Under this assumption, the physical law of heat conduction can be  expressed with a second-order PDE Laplace equation. The two-dimensional form is defined as:

\begin{equation}\label{con:mut2}
\frac{\partial^2T}{\partial x^2}+\frac{\partial^2T}{\partial y^2}=0
\end{equation}

Generally, In the physics-driven method, the Laplace equation is used to drive the training process as the loss function:

\begin{equation}\label{con:mut}
L_{PHY}=\frac{\partial^2T}{\partial x^2}+\frac{\partial^2T}{\partial y^2}=e
\end{equation}

The Dirichlet boundary conditions are applied in the temperature field, where the temperatures of the outer and inner of the plate are constant as zero degrees.

As proposed previously for simple PDEs \cite{sharma2018weakly}, the Laplace equation can be calculated through finite differences using a suitable convolutional filters. Via convolutions, the backpropagation of the loss can be achieved. The weights of the first and second order differenetial kernels are represented as:

\begin{equation}\label{con:mut}
\rm{W}_{\frac{\partial}{\partial x}}
	={\left[ \begin{array}{ccc}
			0 & -0.5 & 0\\
			0 & 0 & 0\\
			0 & 0.5 & 0
		\end{array} 
		\right]},
	\rm{W}_{\frac{\partial^2}{\partial x^2}}
	={\left[ \begin{array}{ccc}
			0 & 1 & 0\\
			0 & -2 & 0\\
			0 & 1 & 0
		\end{array} 
		\right]}.
\end{equation}

\begin{figure}[!h]
	\centering
	\includegraphics[width=1\textwidth]{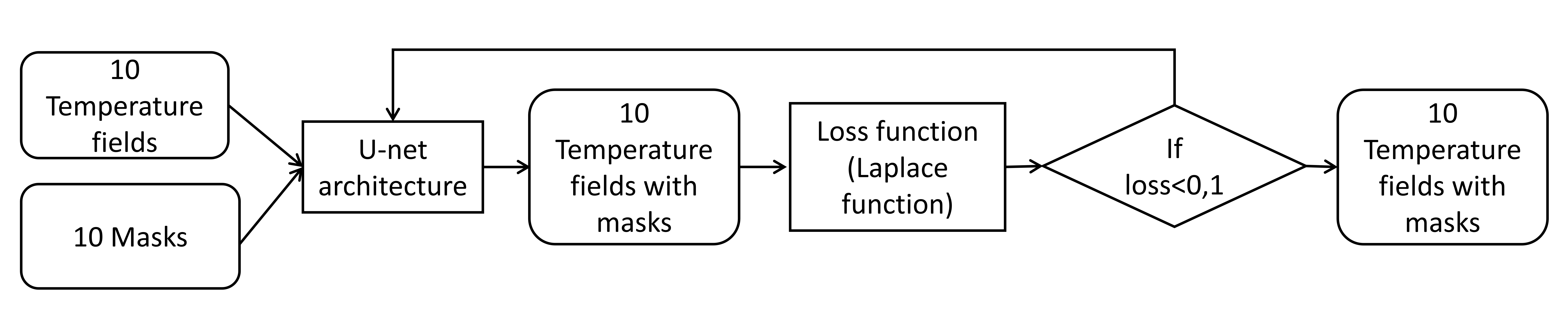}
	\caption{The physics-driven training approach.
	One batch has 10 layout masks.}
	\label{figs:train process}
\end{figure}

The multiple cases training is shown in \Cref{figs:train process}, 10 temperature fields and 10 masks are the two inputs of the U-net architecture. Then the backpropagation process calculates the gradient of the loss function and updates the weights of the multilayer CNN to satisfy the loss function, which is the redisual of Laplace equation. 
At last, when the threshold of loss function is reached, the accurate field solutions are generated.

\subsection{PSO algorithm} 

The \emph{Particle Swarm Optimization} (PSO) algorithm is inspired by groups of birds, in which the basic idea is using collaboration and information sharing between individual birds in the group to find the optimal solution \cite{kennedy1995particle}. 
The feasibility of this method has been proved in many scientific and industrial fields \cite{sterling2011review,kennedy2006swarm,shi2001particle}. 
Similar with the other optimization algorithms \cite{sigmund200199,bendsoe1988generating,mlejnek1993engineer,huang2010evolutionary}, PSO requires a solution space which is conventionally achieved by the discrete methods.  
And when considering complex optimization goals or there are multiple boundary conditions and multiple inputs, the effort of accompanying numerical simulations is usually prohibitive \cite{mueller2019adjoint}.
In this paper, the PD-CNN surrogate model provides an effective way of calculating the solutions inside the domain. 
Given a precised layout and boundary condition, the model is able to directly output corresponding temperature distribution, which significantly reduces the computational workload . 
The PD-CNN mainly replaces the numerical analysis tool by constructing a  high-precision model and has achieved a compromise between calculation accuracy and computational cost, thereby improving the optimization's efficiency. 
It is straightforward to guide the numerical algorithm to solve optimization problem in a limited time. 

\begin{figure}[!h]
	\centering
	\includegraphics[width=0.5\textwidth]{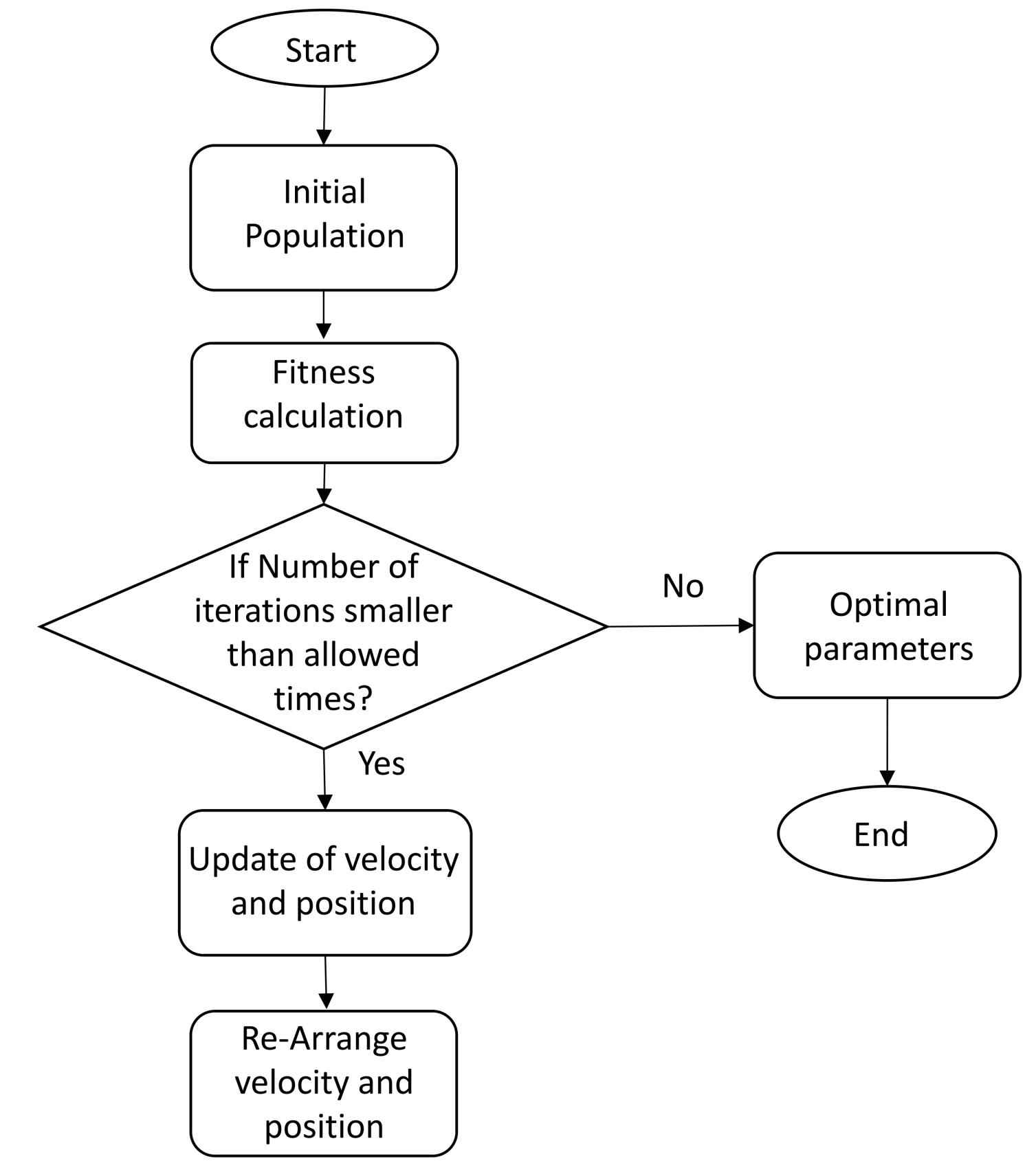}
	\caption{Schematic of the PSO method}
	\label{PSO}
\end{figure}

The PSO process is shown in \Cref{PSO}. 
As introduced above, the birds involves the placement of several simple things, called also particles. In the space of a problem or function, with each particle evaluats the fitness at its present location. By combining some aspect of the history of the swarm's fitness values with those of one or more swarm members, each particle can determine its movement through the parameter space. Then by the locations and processed fitness values of those other members, along with some random perturbations, the swarm can move through the parameter space with a velocity determined. 
The swarm members can interact with its neighborhoods and the colonial neighborhoods of all particles in a PSOs network.

The objective of optimization is to minimize the mean temperature value through the whole precised domain. 
The machine learning methods are able to predict the temperature distribution of multiple cases simultaneously. 
So, in every epoch, ten initial temperature fields with ten different layouts are input into the U-net structure. 
The restriction condition is that the total area of the hole. 
There are three different study cases as shown in \Cref{table:1}:

\begin{table}[!h]
\centering
\caption{Study cases}
\label{table:1}
\begin{tabular}{cccc}
\toprule
Case No. & Number of holes & Length of sides & Position \\
\midrule
case 1      & 1           & 5 $\sim$ 80          & (64,64)      \\
case 2      & 2           & $20 \times 10$       & Free to move    \\
case 3      & 4           & $10 \times 10$       & Free to move   \\
\bottomrule
\end{tabular}
\end{table}

In case 1, the hole position of the mask is located at the exact center of the entire computational domain and the coordinates are (64,64), and the area of the hole is fixed as 400. 
The side length of the hole can be changed, ranging from 5 to 80. In case 2, the side length of the hole is fixed 20 $\times$ 10, so the area of the hole is 200. 
The center position of each hole can be moved up, down, left and right and the moving size is 1, 5 and 10 pixels in each epoch. 
For case 3, the side length of the hole is fixed 10 $\times$ 10, so the area of the hole is 100. 
The center position of each hole is defined as case 2, which can be moved up, down, left and right and the moving size is 1, 5 and 10 pixels in each epoch.


\section{PD-CNN model training} 
\begin{figure}[!h]
	\centering
	\includegraphics[scale=0.2]{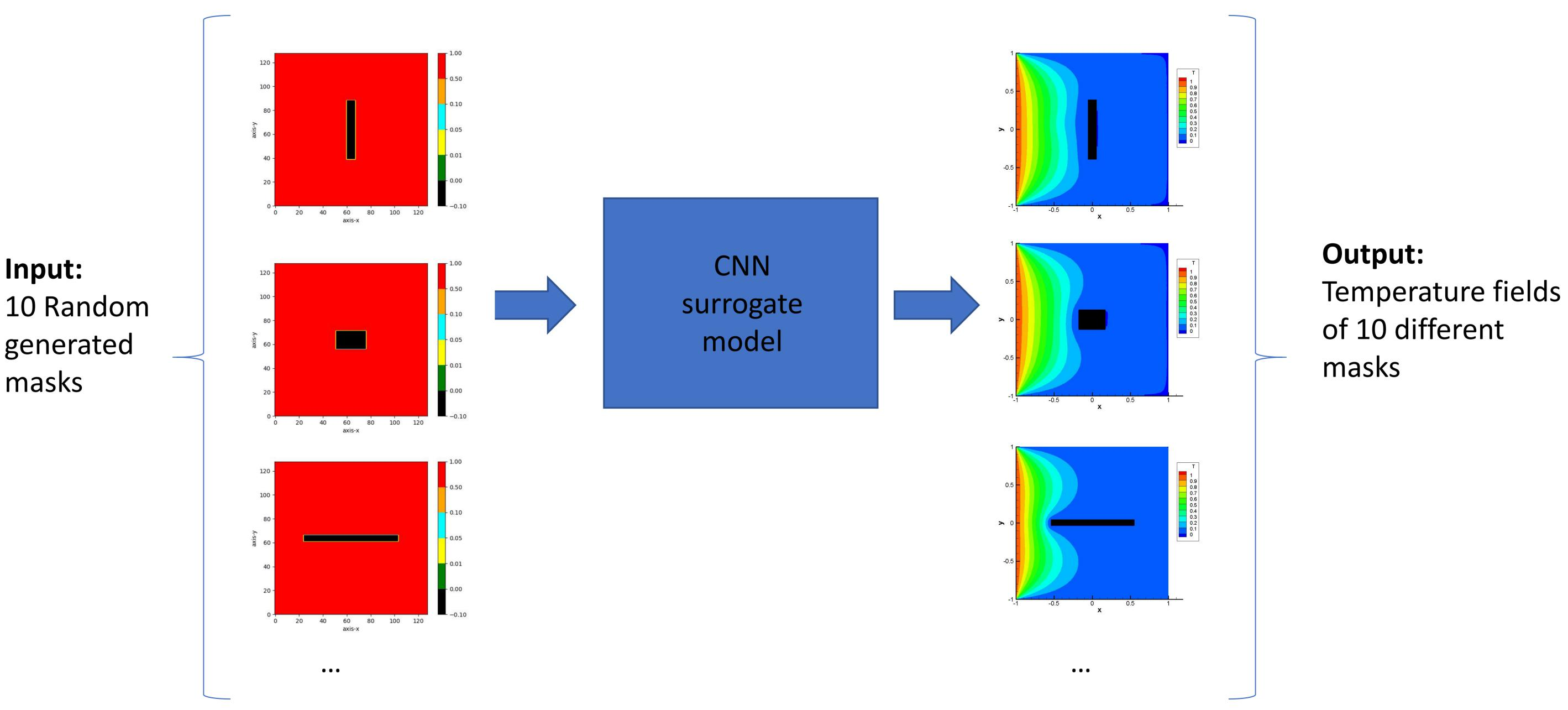}
	\caption{The temperature distribution prediction using the trained PD-CNN surrogate model.}
	\label{figs:CNN surrogate model}
\end{figure}

The CNN surrogate model which is shown in \Cref{figs:CNN surrogate model} is used for the optimization tasks described in the next section. 
The FEM model can provide accurate temperature prediction for the heat transfer, but the optimization based on the FEM model could be infeasible as discussed above. 
The FEM can only provide the solutions with the precised boundary conditions and layouts, so the result is not given in an explicit form. 
PD-CNN model, on the contrary, is actually an accurate mapping between the layouts and corresponding solutions and is able to predict the temperature distribution for any randomly given geometry. 
Therefore CNN surrogate model could be a good choice for layout optimization. 
Figure \ref{figs:CNN surrogate model} shows the input and output of the CNN surrogate model in the case of four holes. 
During every epoch, ten randomly generated masks are used as the input for the CNN surrogate model. 
The masks also change with different epochs. 

\begin{figure}[!h]
	\centering
	
	\subfloat[Iterative\ Steps: 1-500k]{
		\label{convergence plot}
		\includegraphics[width=0.5\textwidth]{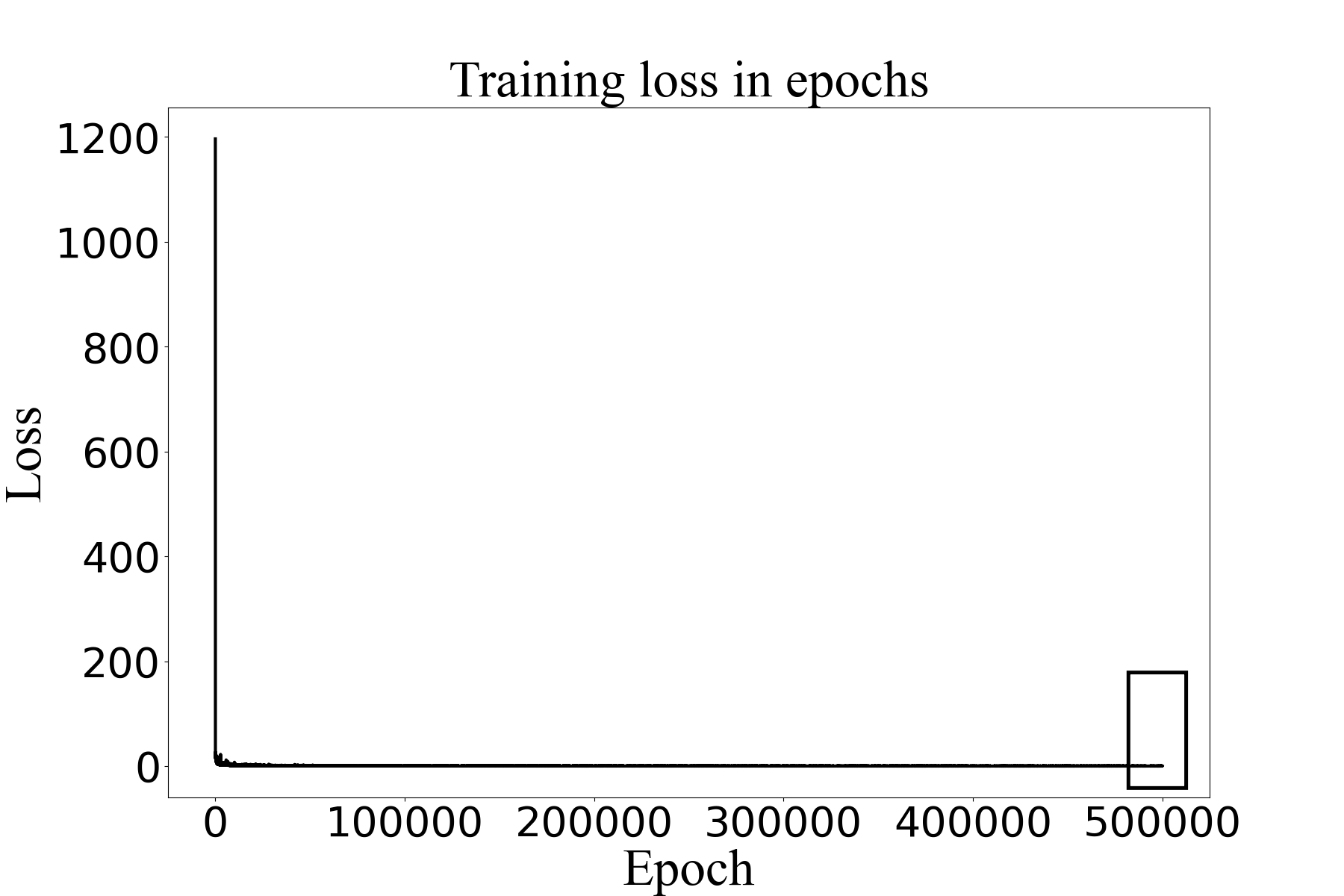}}
	\subfloat[Iterative\ Steps: 499500-500k]{
		\label{convergence plot ZOOM}
		\includegraphics[width=0.5\textwidth]{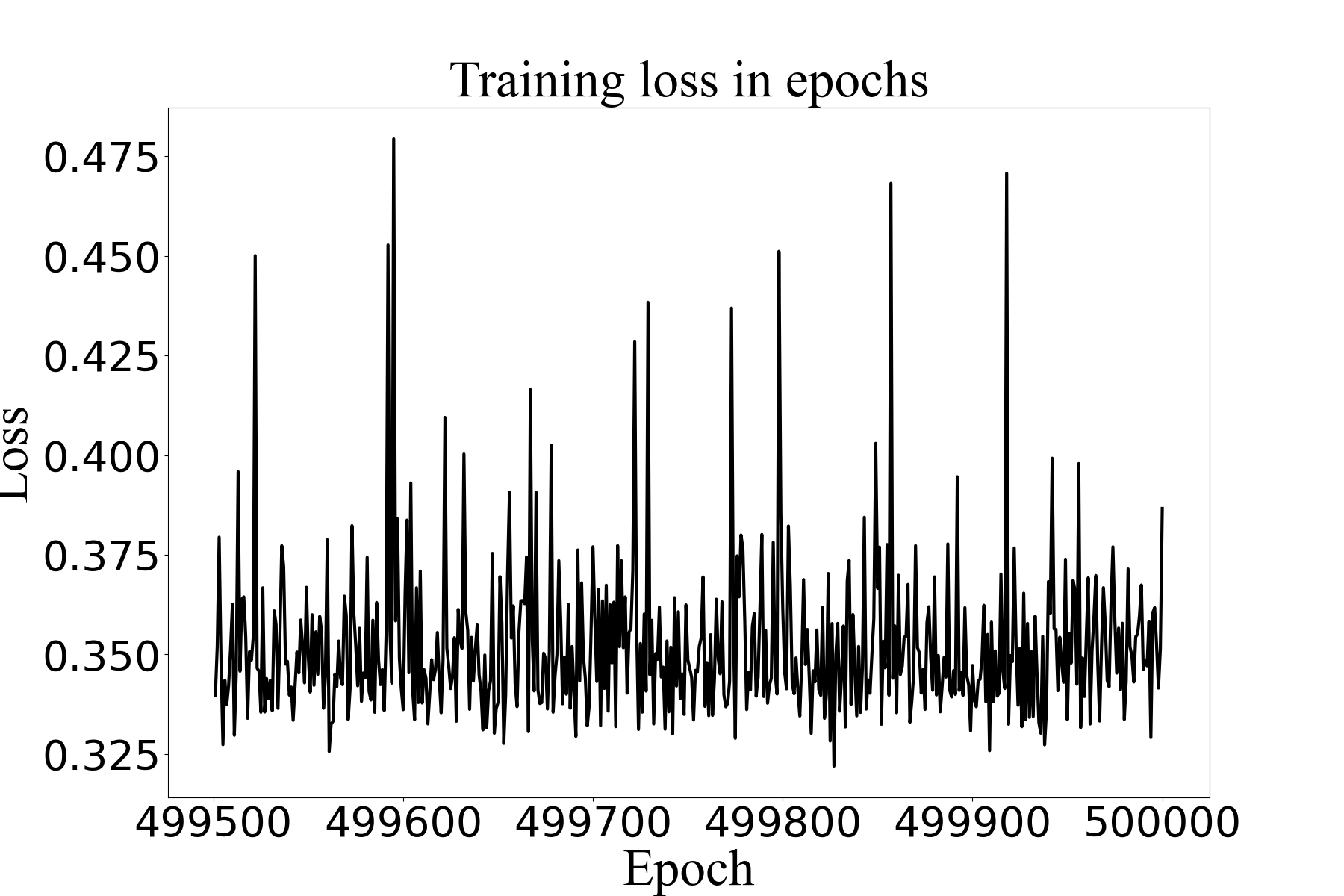}}
	
	\caption{Convergence Curve}
\end{figure} 

As shown in Figure \ref{convergence plot}, the Laplace residual value drops exponentially in the beginning, after approximately 20 iterative steps, the convergence of the Laplace residual exhibits minor differences. 
When the iterative training step reaches 500k, the value decreases to sufficiently small. 
After then, the Laplace residual oscillates around 0.35 as shown in Figure \ref{convergence plot ZOOM}, which is considered converged. 
The whole converge curve is very similar with the curve using the combined data- and physic-driven method in Ref.  \cite{ma2020combined}. 
The whole training process with 500 thousands epochs is needed round 9 hours using the follow device: Intel Core i5-10300H CPU and NVIDIA GeForce RTX 2060.

\begin{figure}[!h]
	\centering

	\subfloat[$Epoch=1$]{
		\includegraphics[width=0.33\textwidth]{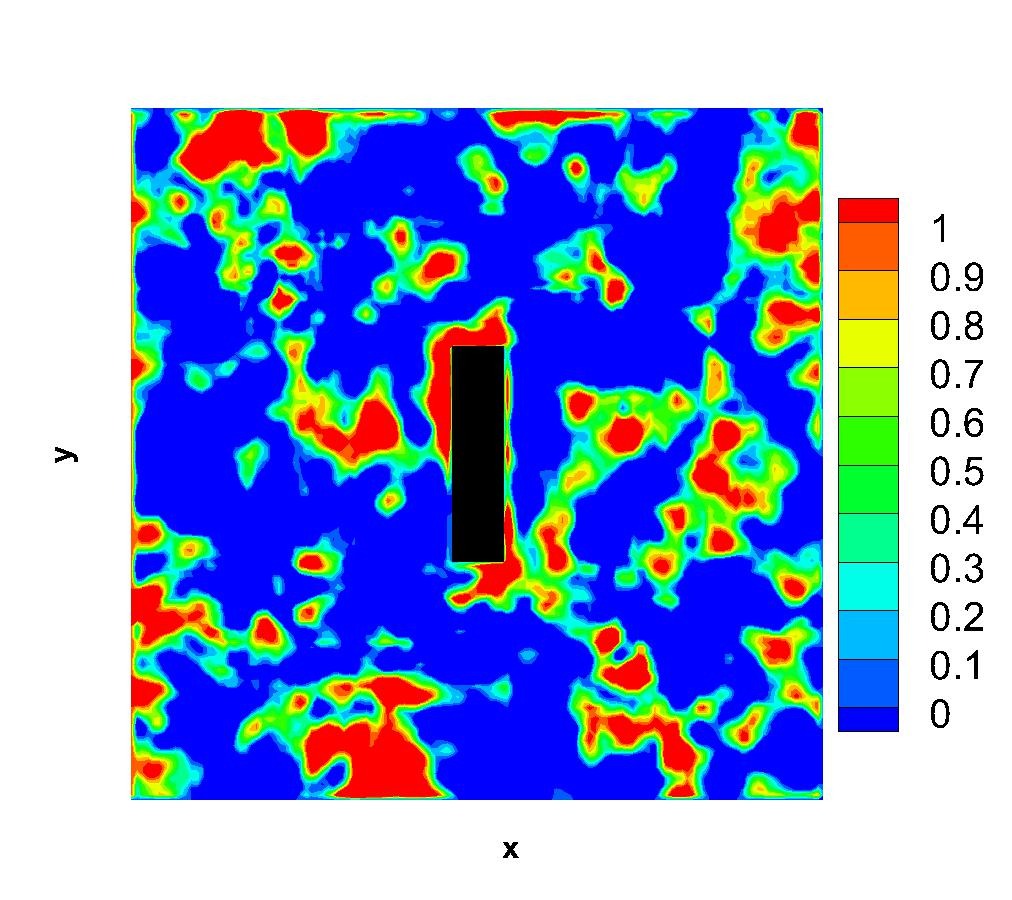}}
	\subfloat[$Epoch=1000$]{
		\includegraphics[width=0.33\textwidth]{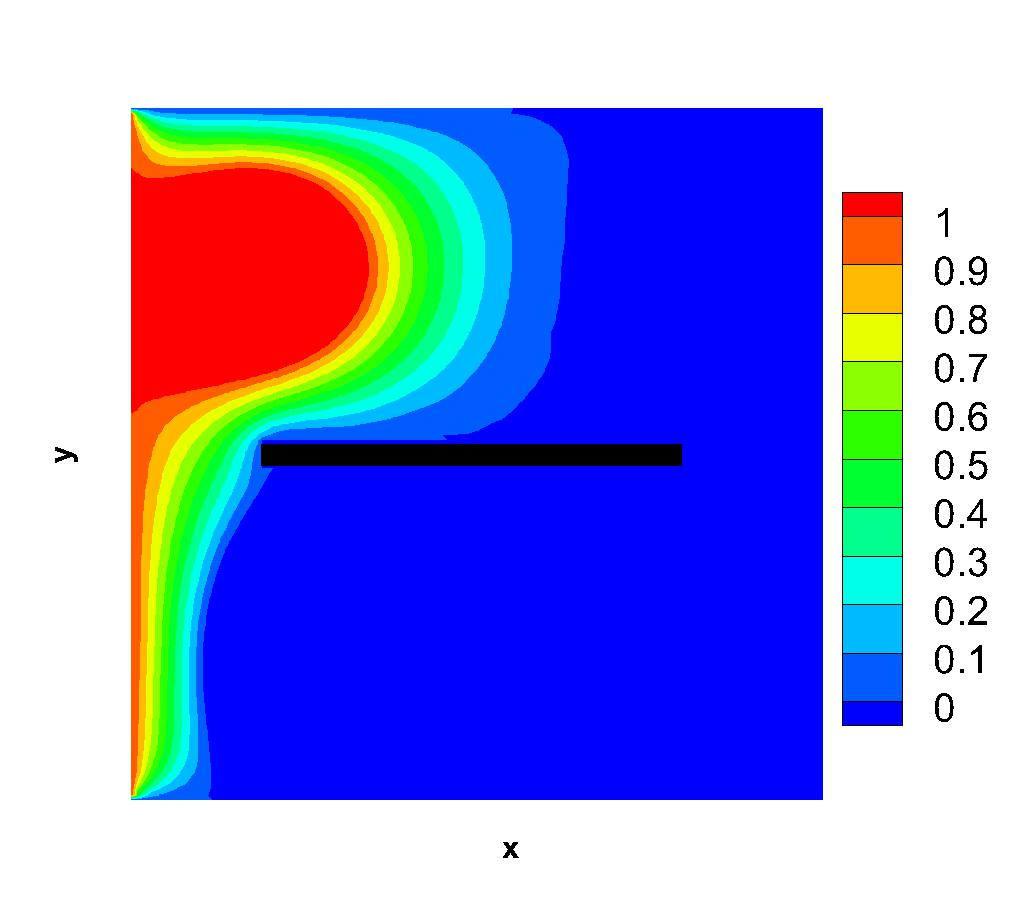}}
	\subfloat[$Epoch=5000$]{
		\includegraphics[width=0.33\textwidth]{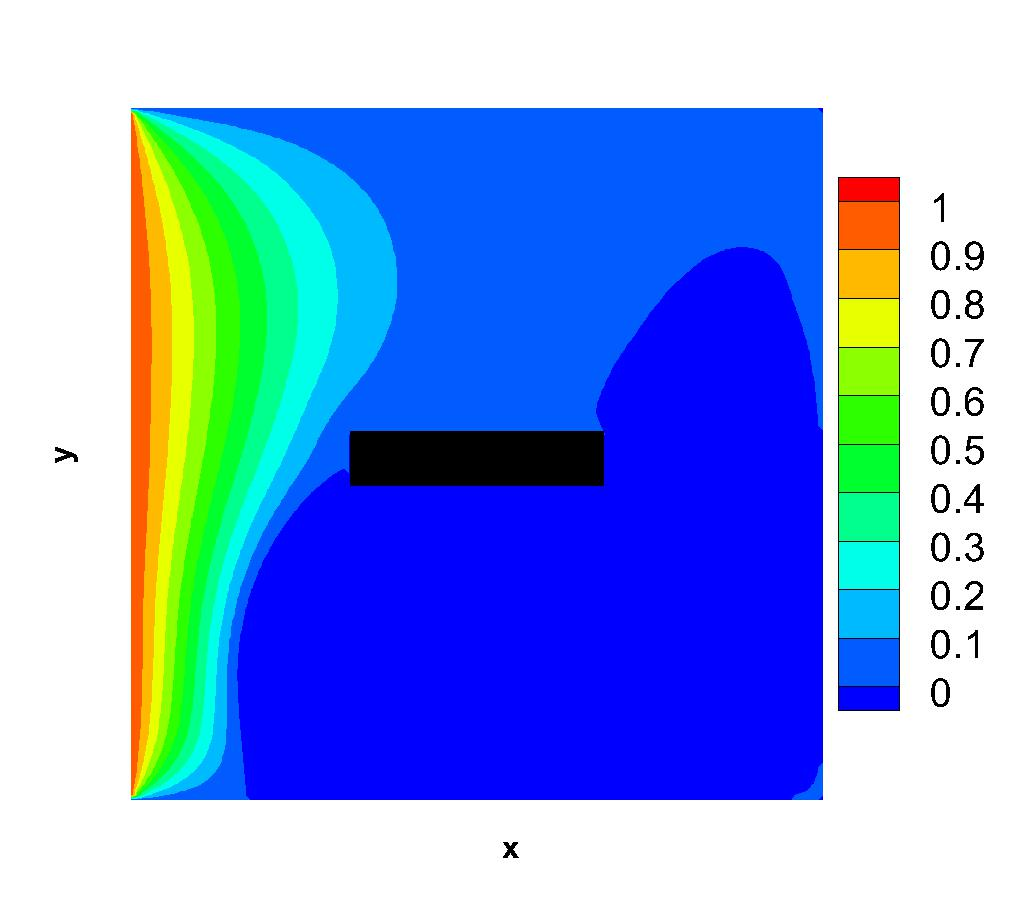}}
	\quad
	
	\subfloat[$Epoch=15000$]{
		\includegraphics[width=0.33\textwidth]{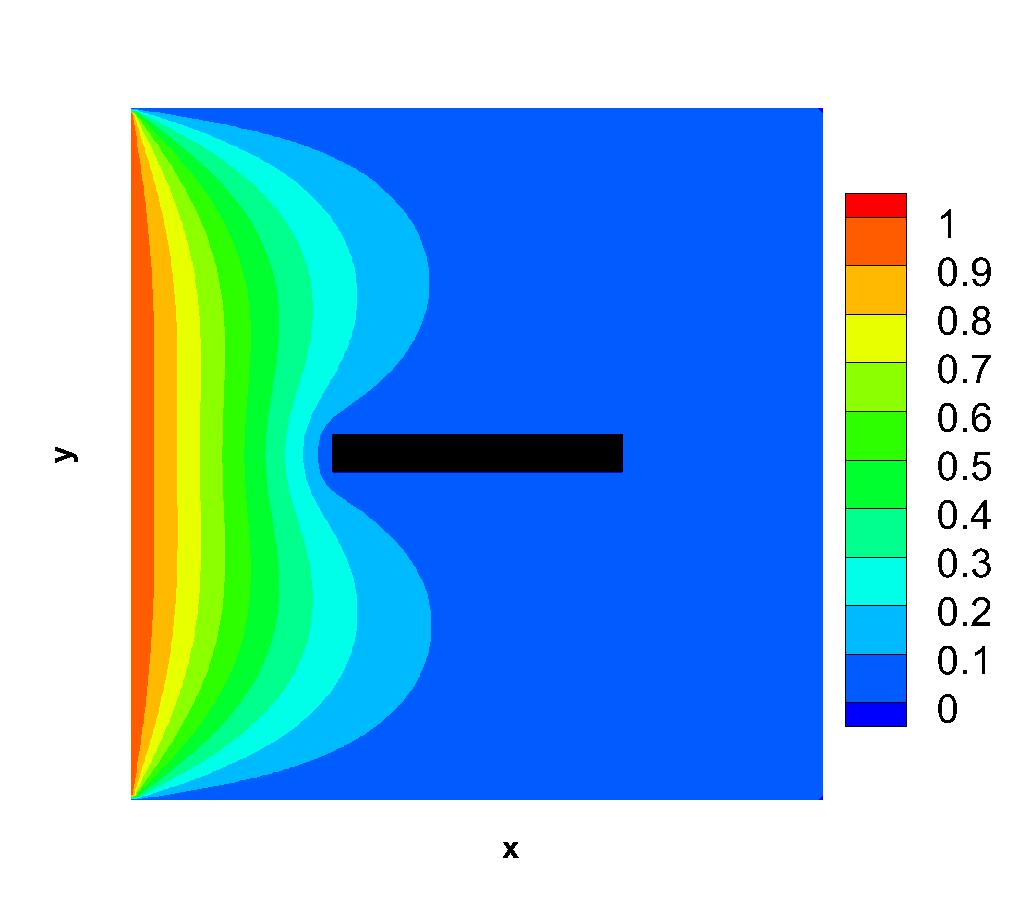}}
	\subfloat[$Epoch=50000$]{
		\includegraphics[width=0.33\textwidth]{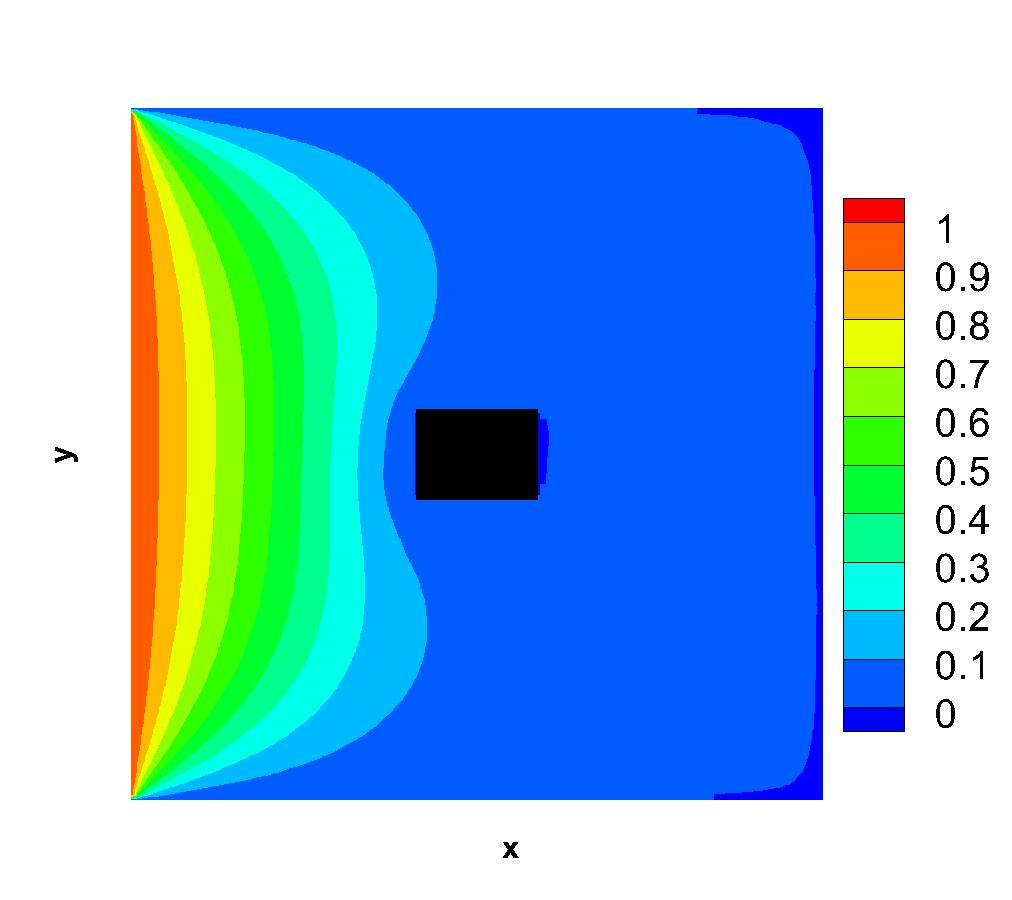}}
	\subfloat[$Epoch=100000$]{
		\includegraphics[width=0.33\textwidth]{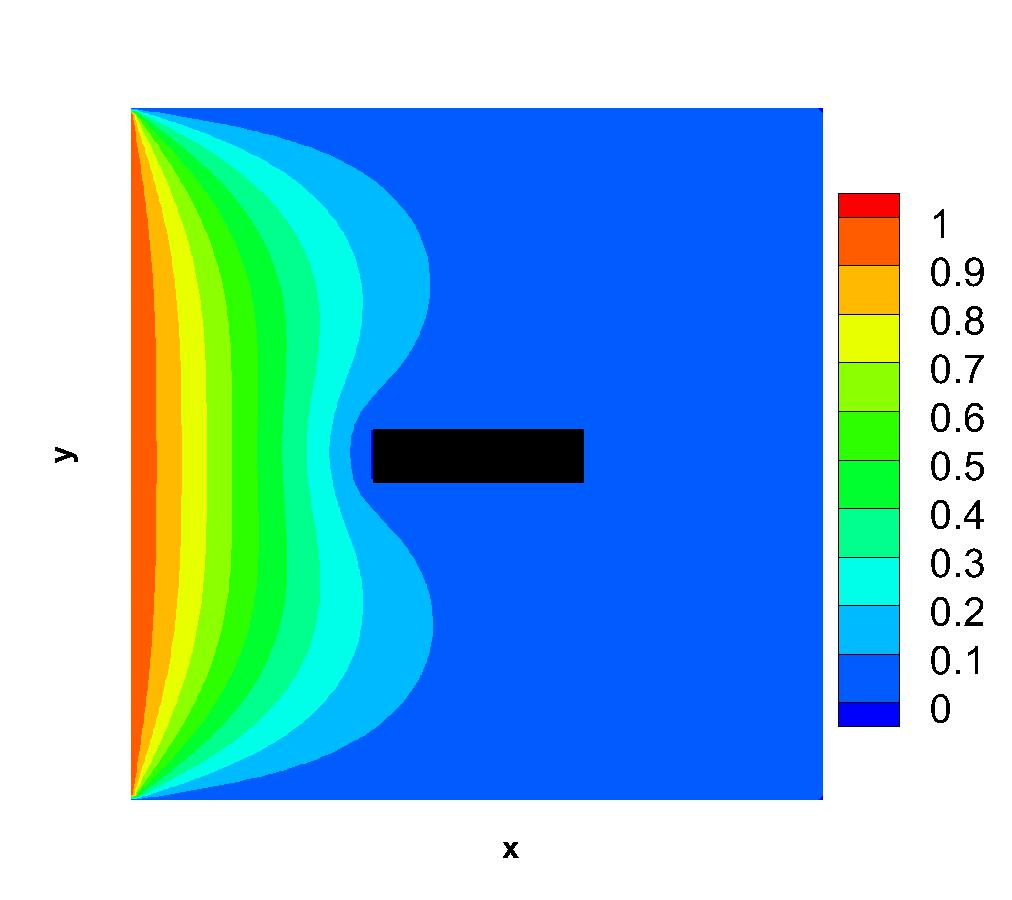}}
	\quad
	
	\subfloat[$Epoch=150000$]{
		\includegraphics[width=0.33\textwidth]{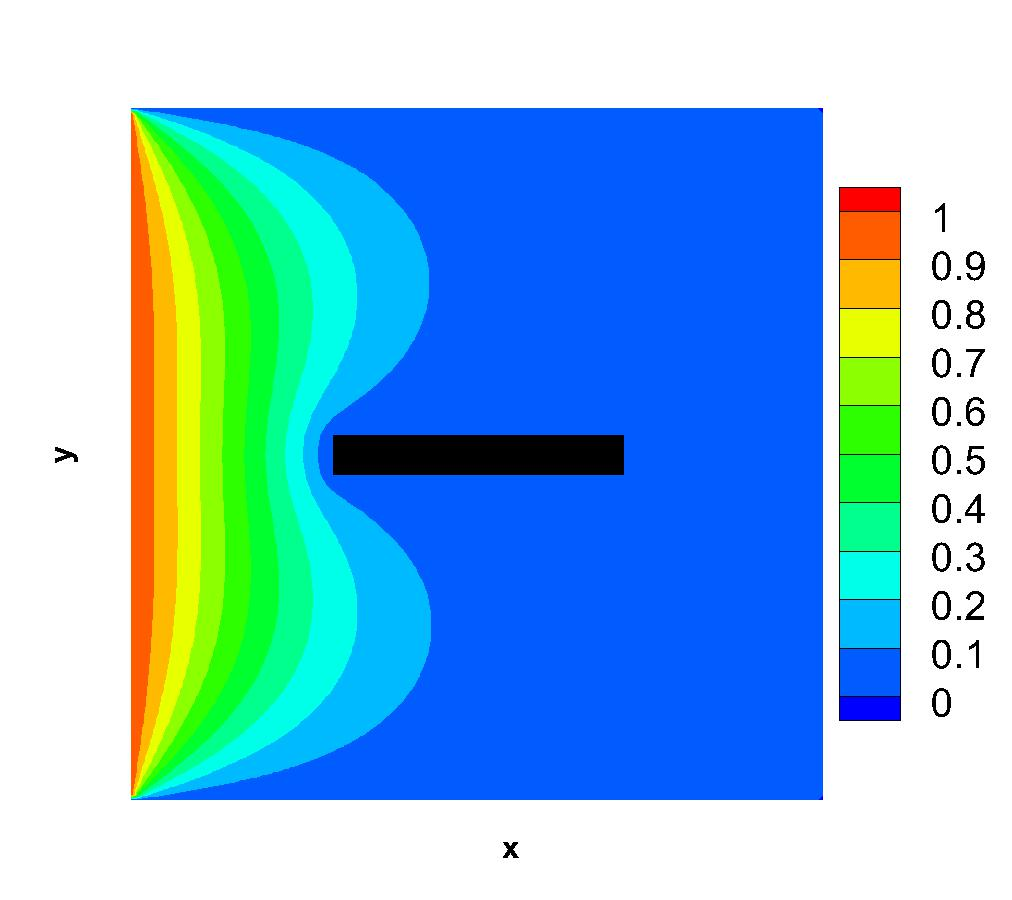}}
	\subfloat[$Epoch=270000$]{
		\includegraphics[width=0.33\textwidth]{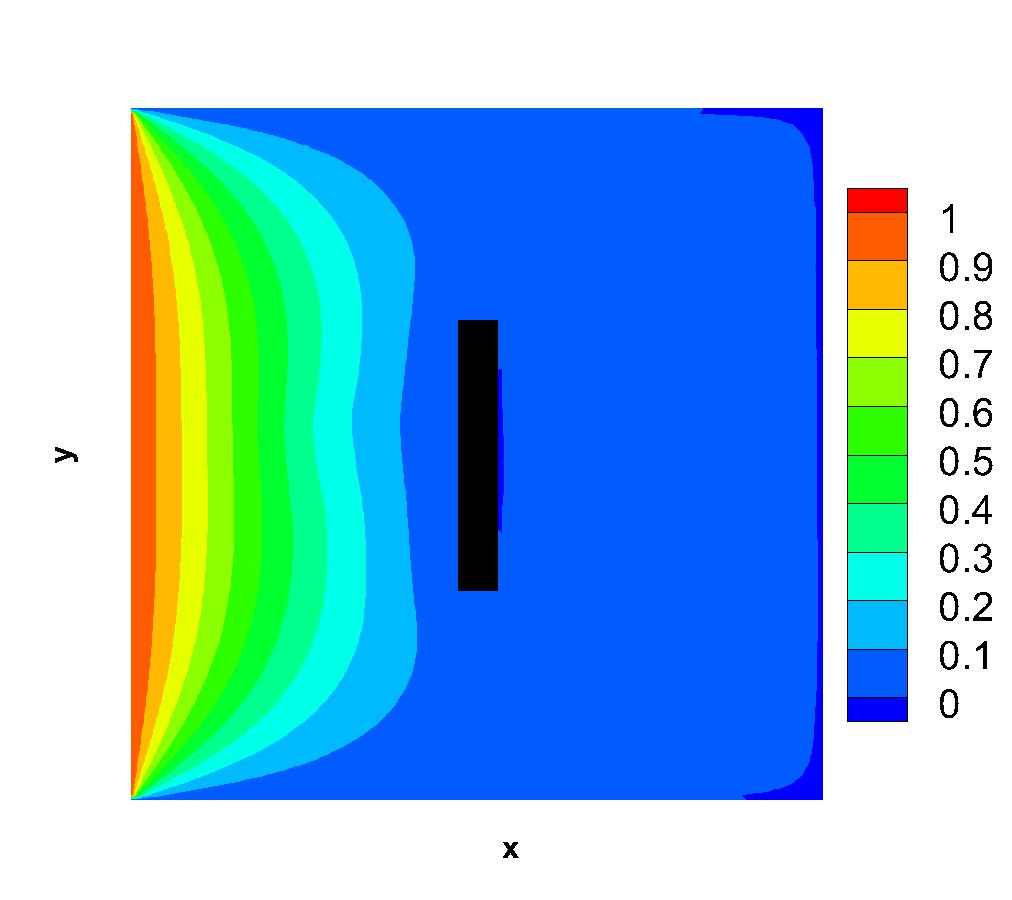}}
	\subfloat[$Epoch=500000$]{
		\includegraphics[width=0.33\textwidth]{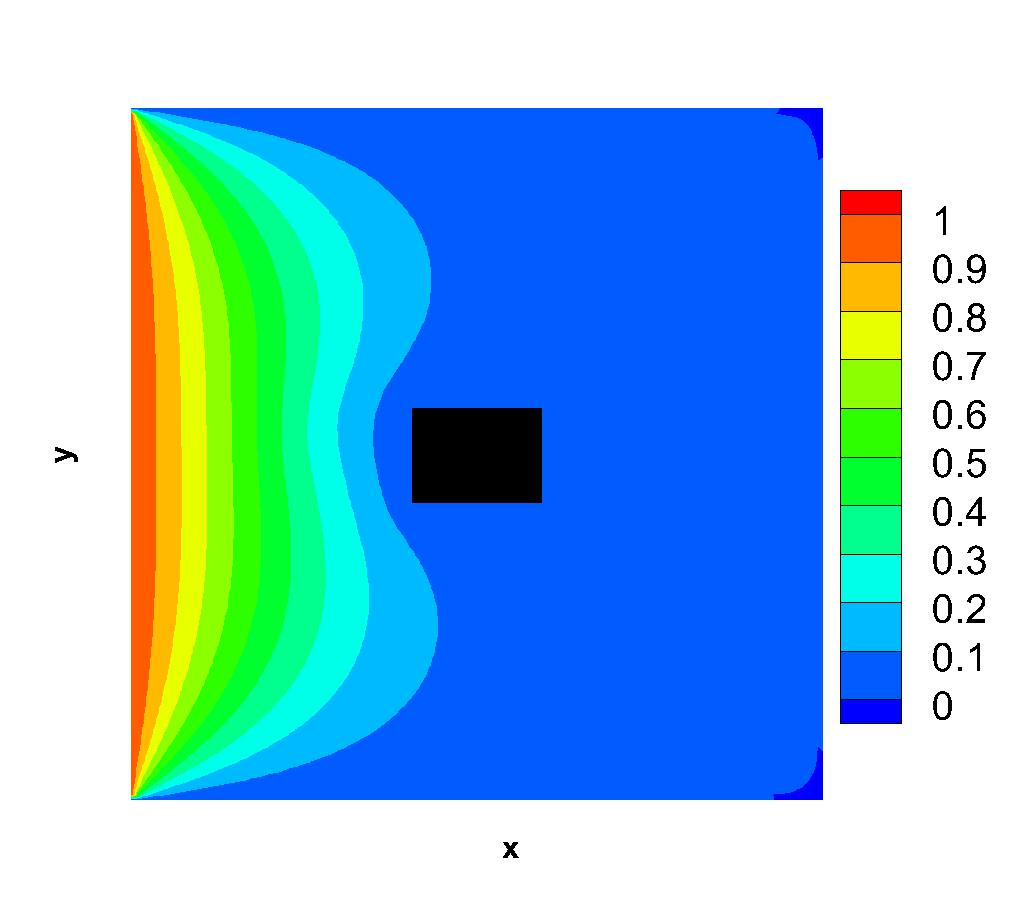}}
	\caption{The training evolution of temperature field solution from epoch 0 to 500000 for the single hole case.}
	\label{train process}
\end{figure}

The training process is shown in \Cref{train process}, 100 thousand are set as the maximal epochs, each epoch the GPU train ten different cases as one batch, and the learning rate is set as 0.001. 
The positions of masks in each epoch is randomly generated. 
The PD-CNN is used to obtain the solutions of multiple cases with a unique surrogate model. 
The input masks are varied in a moderate range in the training stage, the fixed network is able to generate the corresponding temperature field. 
The setup of the mask position is shown in Table \ref{table:1}, as introduced before. 
In the beginning of the training process, training loss is relatively significant, and there are many errors in local temperature fields. 
These errors of inconsistency show big Laplace loss, which cannot predict the temperature field very well. 
As the epochs increase, the temperature field's inconsistency decreases, and the results seem to be more reasonable, as shown in \Cref{train process}. 
The training loss decreases very fast at first and then keeps stable. 
Although the loss does not decrease, the results keep becoming better after 10 thousand epochs. 
In the end, the model adapts to the random size of different masks, which cover the range of the surrogate model completely.

As shown in \Cref{train process}, the mask changes in every epoch  in the learning process, and the contours of the temperature field become smoother. 
As the advantage of physics-driven learning, the relation of adjacent points is constrained by the Laplace equation. For this reason, the values do not vary dramatically. 
Then in epoch 1000, a significant error spot appears, denoted as a peak which the value is round 3, see \Cref{train process}(b), and the residuals near the boundaries of the domain are negative. 
The whole temperature field is much different from the true solution. 
Finally, from epoch 15000, the global structure is stable and very similar to the true solution after the gradual disappearance of the peaks. 
However, there are still residuals near the boundaries of holes, which is a typical numerical solution process of an unsteady heat conduction problem with a Dirichlet boundary condition \cite{ma2020combined}. 
In the validation phase, the surrogate model has the capability to predict the temperature field immediately. 
As the results presented in the next section, the prediction of the PD-CNN surrogate model are almost identical with numerical simulation results.

\section{Prediction and Optimization}
In this section, the prediction results of the temperature field with single hole mask and multiply holes mask are shown. Then the optimization results are shown and discussed in detail. Finally, the results from PD-CNN method are compared with FEM simulation results.

\subsection{Shape optimization in the Single Hole Case}
Here, the single mask dimension is optimized with the PSO method, and the corresponding temperature field is shown and discussed.

\begin{figure}[!h]
	\centering
	\subfloat[PD-CNN]{
		\label{T1 PD-CNN}
		\includegraphics[width=0.33\textwidth]{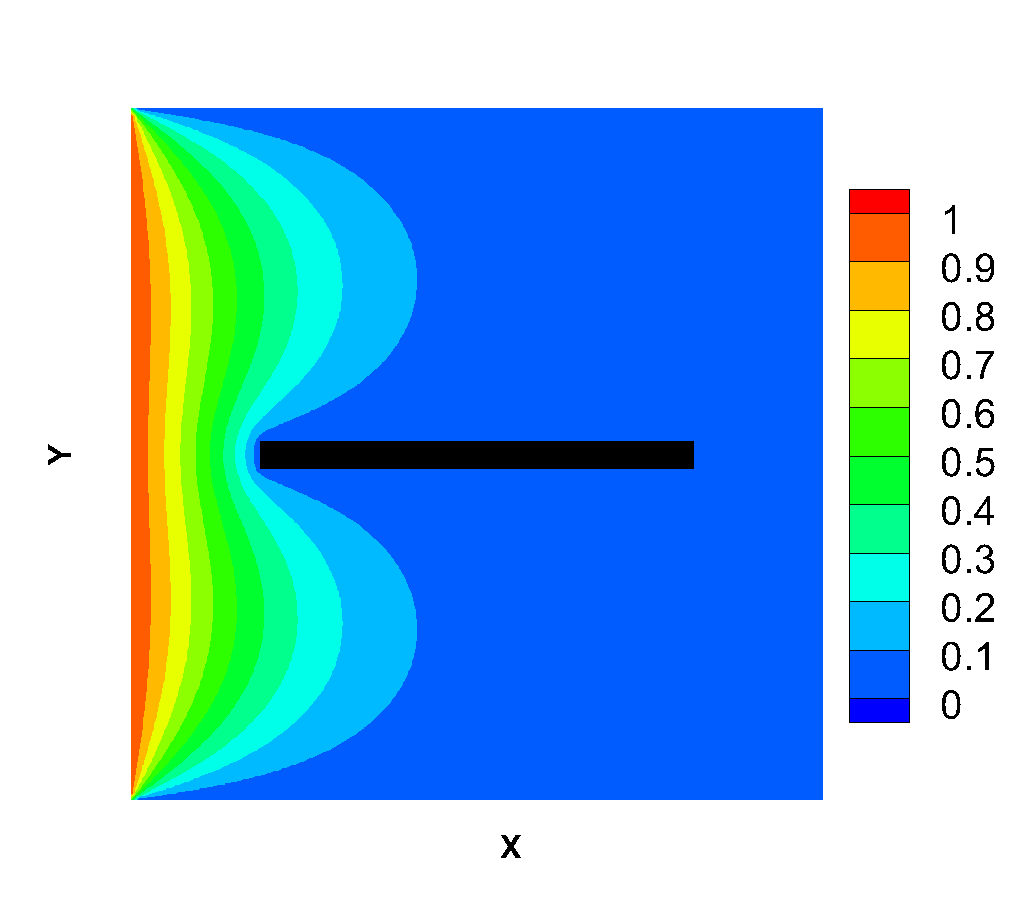}}
	\subfloat[FEM]{
		\label{T1 FEM}
		\includegraphics[width=0.33\textwidth]{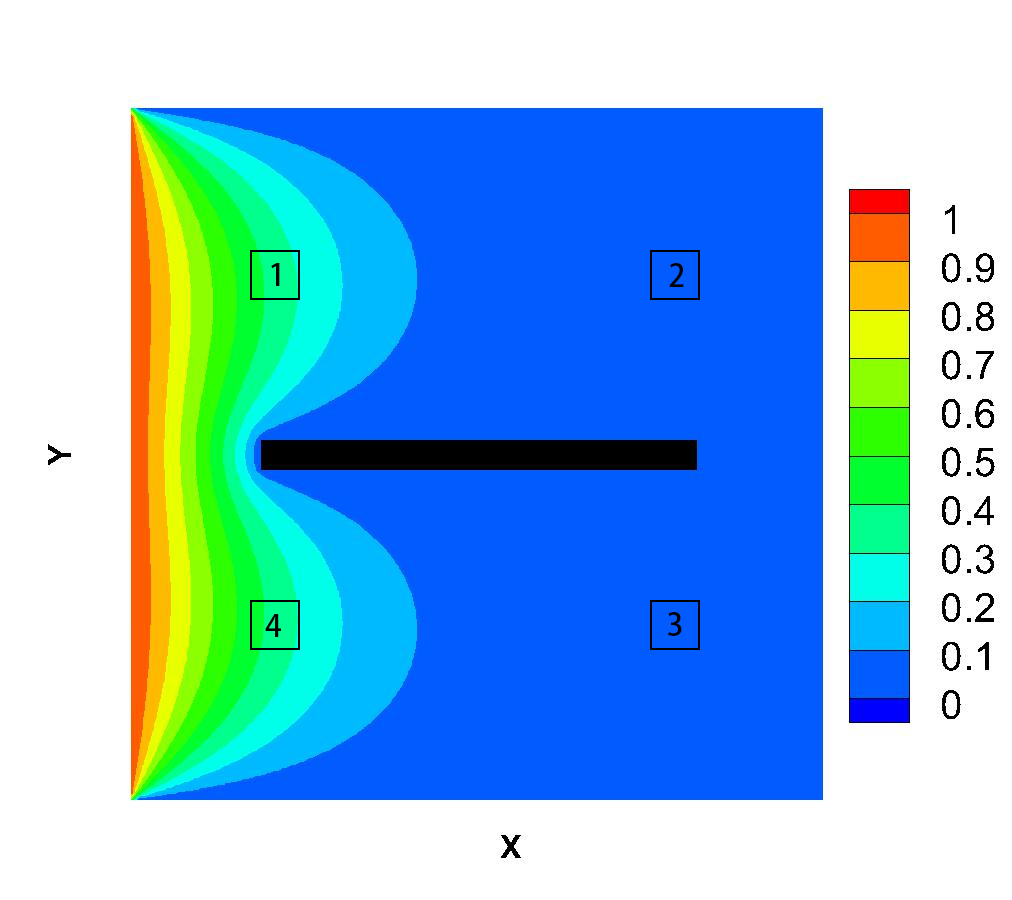}}
	\subfloat[PD-CNN vs FEM]{
		\label{comparison1}
		\includegraphics[width=0.33\textwidth]{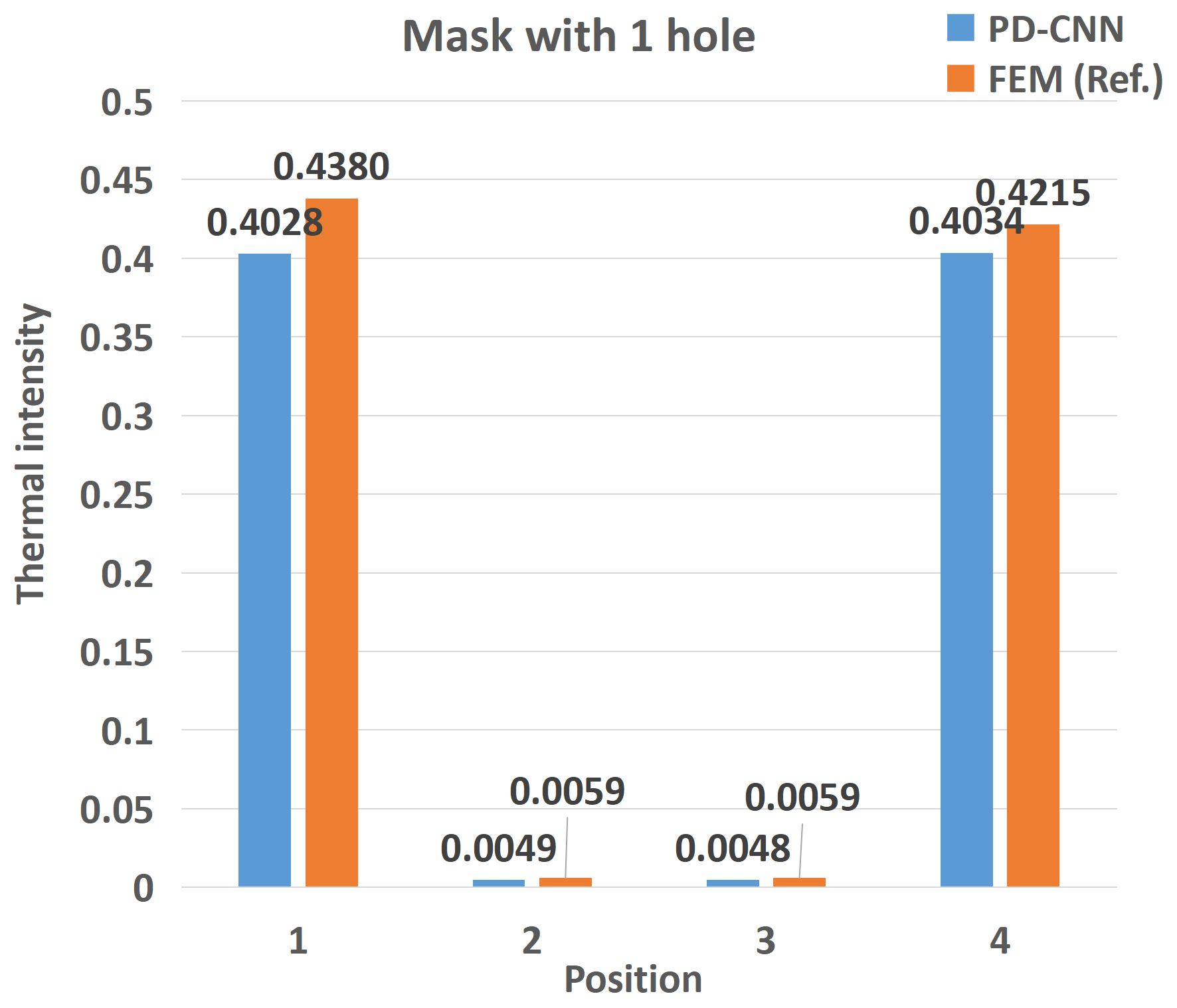}}
	\quad	
	\caption{Comparision between surrogate model prediction and FEM result for 1 hole case. 
		Left:
		temperature field with optimized mask using CNN surrogate model; Middle:
		temperature field with optimized mask using FEM method; Right: quantitative comparison of the two results.}
\end{figure} 

\begin{table}[!h]
\caption{The shape optimization of the 1 hole case.}
\begin{tabular}{ccc}
\toprule
Hole No. & Side length before optimization & Side length after optimization \\
\midrule
1th hole       & $20 \times 20$                  & $5 \times 80$   \\
\bottomrule           
\end{tabular}
\end{table}

The area of every hole is fixed as 400, and every side length of the hole is variable in the range from 5 to 80. 
The mask is at the central position of the whole domain. 
After the PSO optimization process using a PD-CNN surrogate model, the  temperature field with teh optimized mask is shown in \Cref{T1 PD-CNN}. 
The width of the optimized mask is 5, and the height is 80. 
The mask has divided the temperature field into upper and lower two parts. 
According to Table 2, the mask's width should be as small as possible to reduce the heat transfer from left to right. 
With this optimized mask, the mean temperature of the whole area is minimum. 
The \Cref{T1 FEM} is the result from FEM, which is regarded as the true reference, the result of PD-CNN has a good agreement with it.

For a quantitative analysis, four square sampling areas are marked in the temperature field obtained by the two methods. 
The length of each side of the square is 5, and the distance from the center point to the boundary is 20. 
The temperature values in each square are averaged, so that the difference between two results can be intuitively observed and compared, as shown in \Cref{comparison1}. 
It can be seen from the histogram that the difference between the results obtained by the PD-CNN method and the FEM method in sampling areas 1 and 4 is larger than that in sampling areas 2 and 3. 
The possible reason is that the heat source is located on the left boundary, the temperature is transmitted from left to right, the value changes greatly in the area close to the left, and the calculation results are slower to converge, which leads to larger errors. 
The results of sampling areas 2 and 3 on the right The difference between the numerical value and the reference numerical value is small. 
The reason is that the heat source propagation is blocked due to the blocking of the hole, so that the numerical value on the right side of the hole changes less, the calculation converges more quickly, and the difference is smaller.


\subsection{Position optimization in the Multiple Holes Cases}
For the temperature field with two masks, the area of each is fixed as $20 \times 10$, and the central position of each hole is movable from the center of whole domain, which is $(64, 64)$. the movement of both coordinate x and y is in the range from -10 to 10. 

\begin{figure}[H]
	\centering
	\subfloat[PD-CNN]{
		\label{T2 PD-CNN}
		\includegraphics[width=0.33\textwidth]{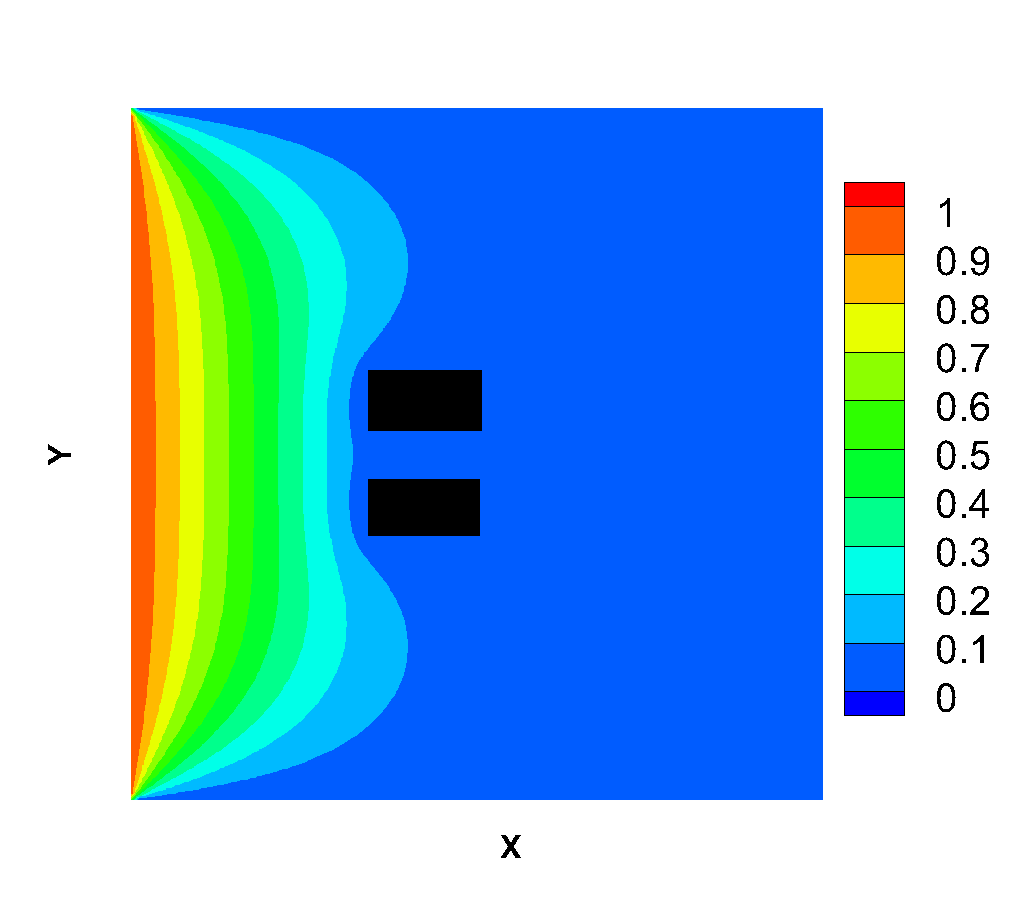}}
	\subfloat[FEM]{
		\label{T2 FEM}
		\includegraphics[width=0.33\textwidth]{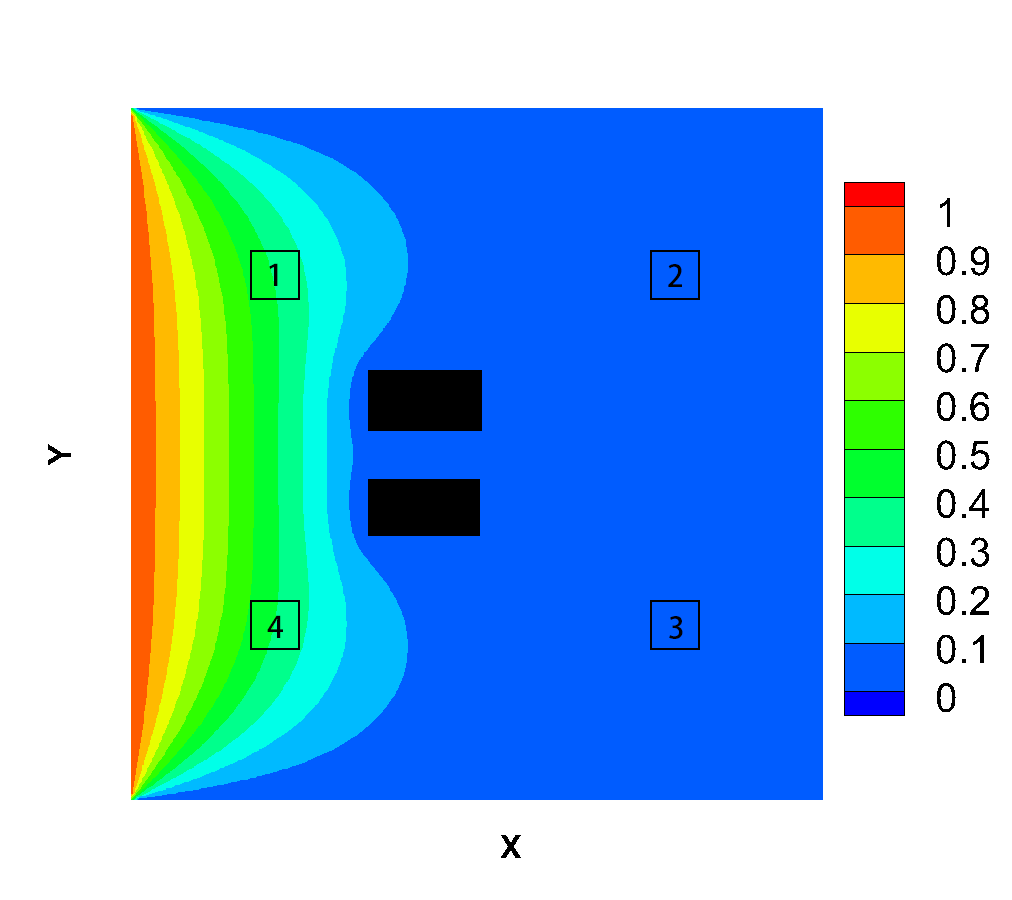}}
	\subfloat[PD-CNN vs FEM]{
		\label{comparison2}
		\includegraphics[width=0.33\textwidth]{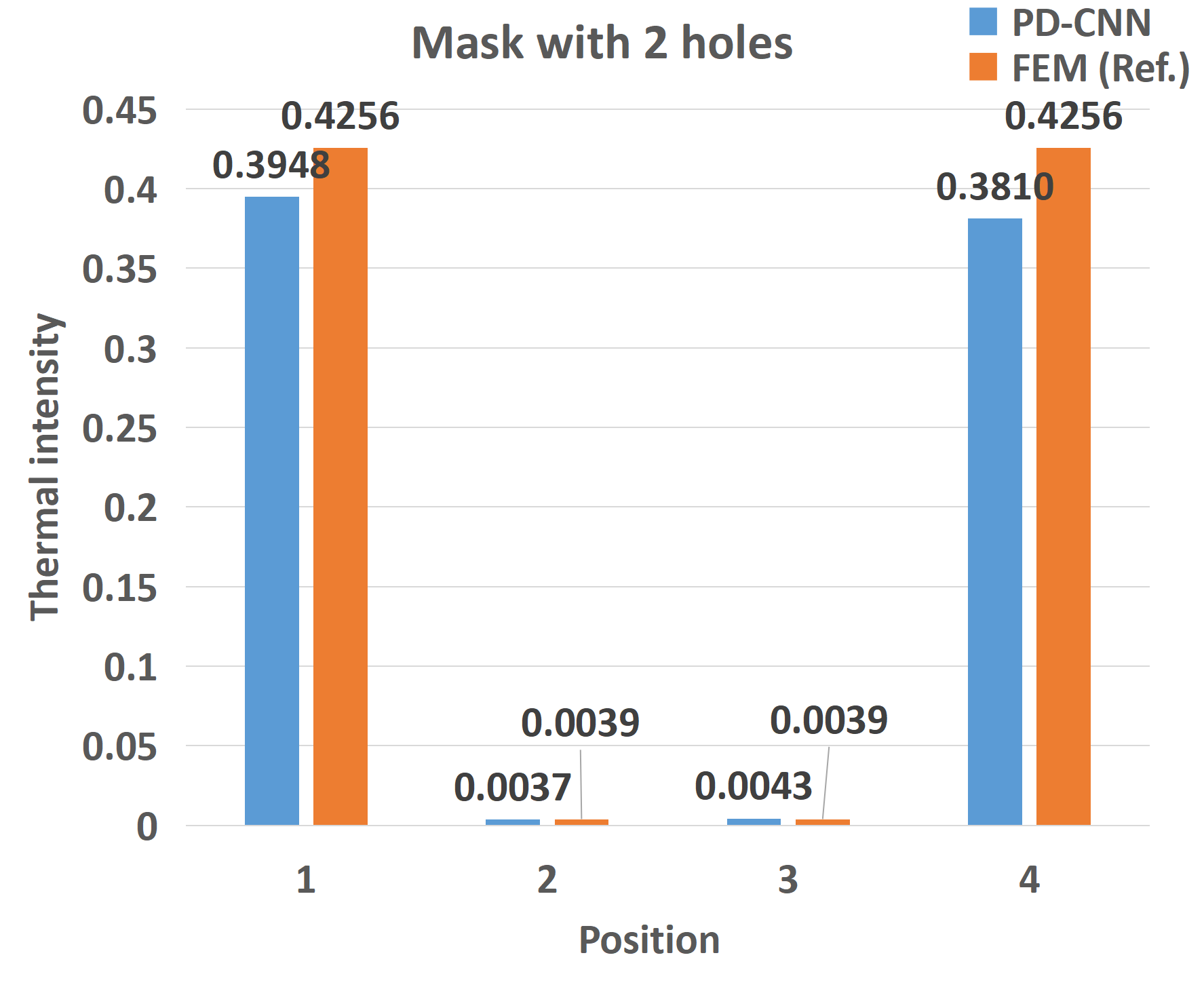}}
	\quad
	\caption{Comparision between surrogate model prediction and FEM result for the 2 holes case.  Left:
	temperature field with optimized mask using CNN surrogate model; Middle:
	temperature field with optimized mask using FEM method; Right: quantitative comparison of the two results.}
\end{figure} 

\begin{table}[H]
\centering
\caption{The position optimization of the 2 holes case. The position is represented by the coordinate of the hole center.}
\begin{tabular}{ccc}
\toprule
Hole No.    & Position before optimization & Position after optimization \\
\midrule
1th hole  & (64, 64)              & (54, 54)            \\
2th hole  & (64, 64)              & (54, 74)			\\
\bottomrule             
\end{tabular}
\end{table}

The Figure \ref{T2 PD-CNN} shows the temperature field with two optimized masks. 
The area of each mask is fixed as 20 $\times$ 10. 
The central coordinate of each mask is movable inside the domain . 
Before optimization, two masks were in the same position, which is (64,64). 
The movement of both coordinate x and y of each mask ranges from -10 to 10. 
Table 3 shows the coordinate of masks after optimization. 
It is noticed that one mask was moved to the left upper part while another was moved to the left lower part. 
In this situation, the square plate has the minimum mean temperature. Again, the \Cref{T2 FEM} is result from FEM, which is defined as the true reference. 
Analyzing from a qualitative perspective, the result from PD-CNN has a good agreement with that of FEM.

In this case, we can observe a similar result compared with 1 hole case, that is, the difference between the sampling area near the heat source and the reference value is greater than the difference between the sampling area on the right side of the mask.


\begin{figure}[!h]
	\centering
	\subfloat[PD-CNN]{
		\label{T4 PD-CNN}
		\includegraphics[width=0.33\textwidth]{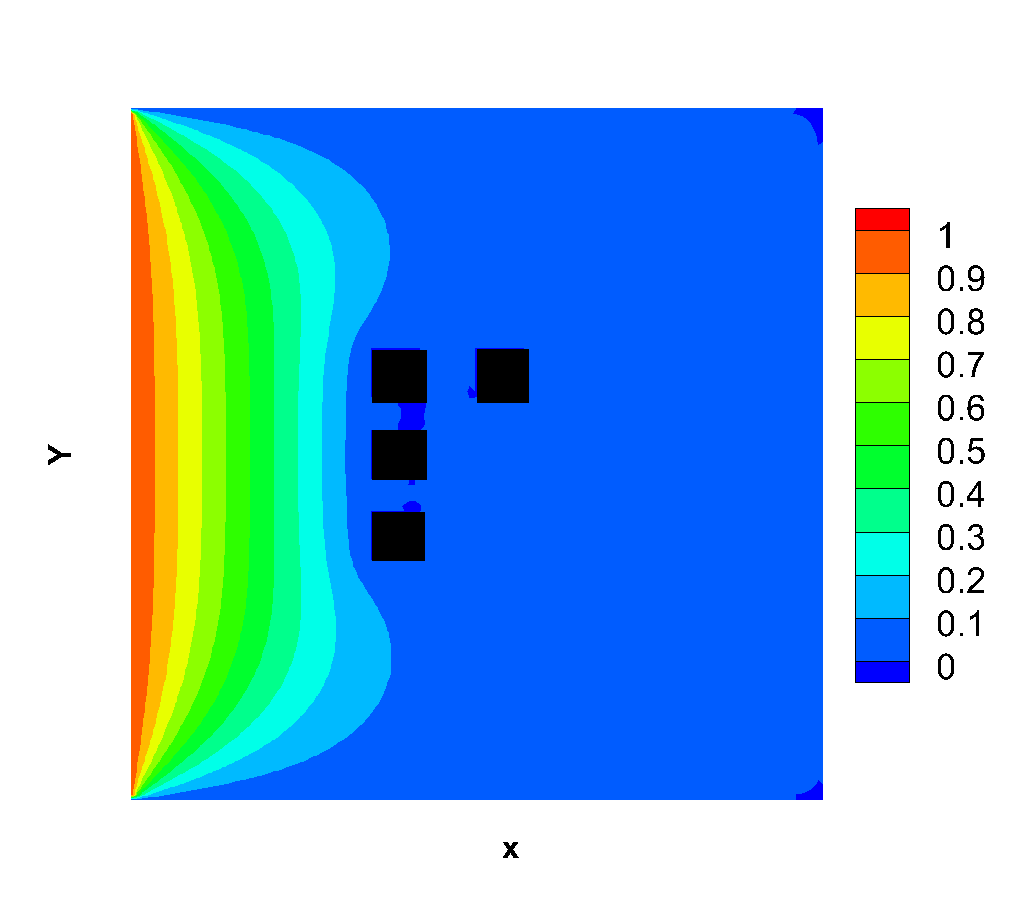}}
	\subfloat[FEM]{
		\label{T4 FEM}
		\includegraphics[width=0.33\textwidth]{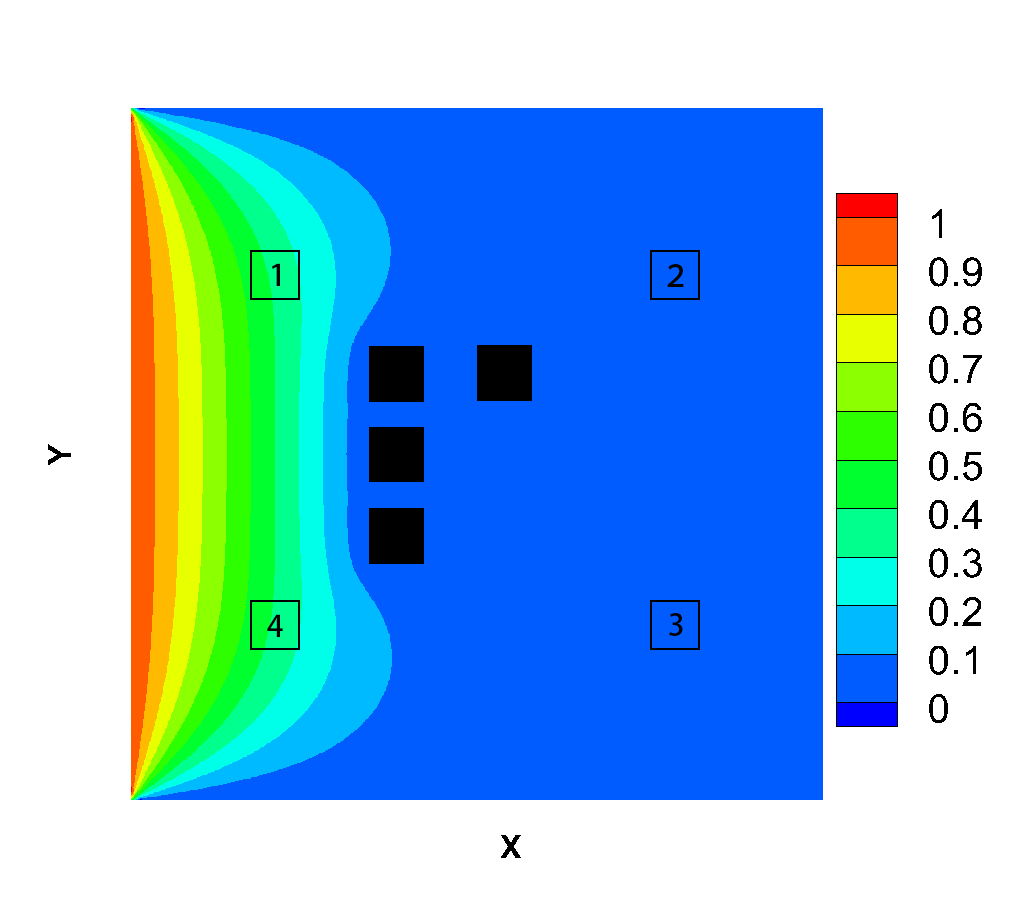}}
	\subfloat[PD-CNN vs FEM]{
		\label{comparison4}
		\includegraphics[width=0.33\textwidth]{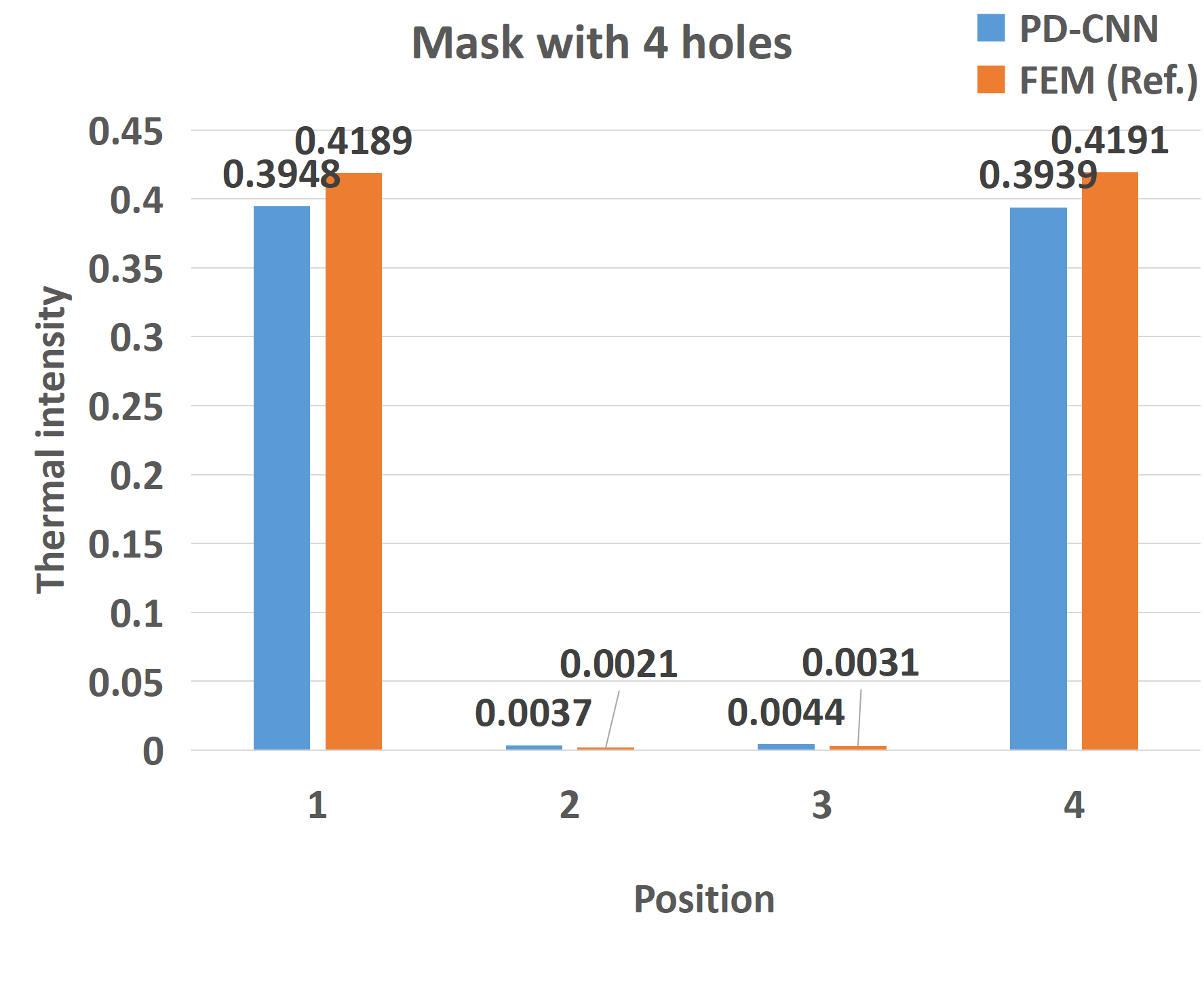}}
	\quad
	\caption{Comparision between surrogate model prediction and FEM result for 4 holes case. Left:
	temperature field with optimized mask using CNN surrogate model; Middle:
	temperature field with optimized mask using FEM method; Right: quantitative comparison of the two results.}
\end{figure}

\begin{table}[!h]
\centering
\caption{The position optimization of the 4 holes case. The position is represented by the coordinate of the hole center.}
\begin{tabular}{ccc}
\toprule
Hole No.          & Position before optimization & Position after optimization \\
\midrule
1th hole        & (64, 64)              & (49, 49)             \\
2th hole      & (64, 64)              & (49, 64)             \\
3th hole       & (64, 64)              & (49, 79)             \\
4th hole		 & (64, 64)              & (69, 79)             \\
\bottomrule
\end{tabular}
\end{table}

The \Cref{T4 PD-CNN} shows the temperature field with four optimized masks using PD-CNN method. 
The area of each hole is fixed as 10 $\times$ 10, and the central position of each hole is movable from the center of the plate. 
Four masks were in the same position, which means they were coincident initially. 
The movement of both coordinate x and y of each mask ranges from -20 to 20. 
We can notice that three holes are arranged vertically in the middle and left part of the entire domain, and the fourth mask is located in the right and upper of the whole domain. 
Considering the symmetry of the computational domain, we can easily infer the distribution of another solution: the top hole is on the right upper side. 
Also in this case, we can see the numerical difference distribution similar to the previous two cases from \Cref{comparison2}, The only difference is that the temperature values obtained by PD-CNN have residuals in the four corners. The reason is that the calculation is more complicated in the case of 4 holes, and the convergence speed will decrease, but as the number of training increases, this residual will gradually disappear.

In the above three cases, the results obtained by the traditional FEM method were used as the references, and the difference between results obtained by the PD-CNN method and the results obtained by the FEM method was compared, which proved from a qualitative point of view the high effectiveness of the new method. 
From a quantitative point of view, the article compares difference in numerical average of the results obtained by the two methods in 4 different specific areas, which verifies the high accuracy of the new method.

\section{Conclusion}
In this paper, we proposed a optimization framework based on the PSO algorithm and PD-CNN surrogate model for the layout of a heat conducting plate.
The Laplace equation is utilized as loss function during the learning process.
So the U-net structure CNN is trained to predict the accurate temperature solution of the precised layouts without any training data. 
The PSO algorithm is used to get the optimized shape and position of the heat insulation holes in the single and multiple holes optimization. 
The solution predictions have a good agreement with the FEM results. 
This optimization framework is very useful for practical application when the training data is not accurate or available. 
It is noteworthy that the U-net architecture using CNN is suitable for general physics laws, which can be expressed as PDEs, so for further research, complex aerodynamic optimization tasks will be carried out.

\section{Acknowledgement}

Hao Ma (No. 201703170250) is supported by China Scholarship Council when he conducted the work this paper represents. 
Yang Sun is supported by Europe Erasmus Exchange Program in the Technical University of Munich (TUM), Germany. 
The author thanks the colleagues of TUM and University of Pisa (UNIPI) for the beneficial discussions.

\clearpage

\bibliography{mybibfile3}

\end{document}